\documentclass[runningheads]{llncs}
\newif\ifready
\readytrue

\usepackage{eccv}

\definecolor{TableBlue}{rgb}{0.17,0.49,0.75}
\definecolor{Cerulean}{rgb}{0,0,0.95}
\definecolor{LimeGreen}{rgb}{0.15,0.65,0.15}
\definecolor{RoyalBlue}{rgb}{0.25,0.41,0.88}
\definecolor{Rose}{rgb}{1.0, 0.15, 0.21}
\definecolor{Orange}{rgb}{1.0, 0.5, 0.0}
\definecolor{Gray}{gray}{0.6}
\definecolor{Black}{gray}{0.0}
\definecolor{Purple}{rgb}{0.77,0.12,0.64}

\usepackage{threeparttable}

\usepackage{mathrsfs}
\usepackage{times}
\usepackage{epsfig}
\usepackage{graphicx}
\usepackage{amsmath}
\usepackage{amssymb}
\usepackage{subcaption} %
\usepackage{comment}
\usepackage{graphicx,scalerel}

\usepackage{eccvabbrv}

\usepackage{graphicx}
\usepackage{booktabs}
 \usepackage{multirow} 
\usepackage[accsupp]{axessibility}  %

\usepackage{hyperref}
\usepackage{orcidlink}
\author{Qingwen Zhang\inst{1}\orcidlink{0000-0002-7882-948X} \and
Xiaomeng Zhu\inst{1}\orcidlink{0000-0002-4180-3809} \and
Chenhan Jiang\inst{2}\orcidlink{0000-0001-8771-3641} \and
Patric Jensfelt\inst{1}\orcidlink{0000-0002-1170-7162}
}

\authorrunning{Q~Zhang et al.}

\institute{KTH Royal Institute of Technology, Stockholm, Sweden \and
Hong Kong University of Science and Technology, Hong Kong, China
}
\begin{document}

\title{SynFlow: Scaling Up LiDAR Scene Flow Estimation with Synthetic Data} 
\maketitle

\begin{abstract}
Reliable 3D dynamic perception requires models that can anticipate motion beyond predefined categories, yet progress is hindered by the scarcity of dense, high-quality motion annotations. 
While self-supervision on unlabeled real data offers a path forward, empirical evidence suggests that scaling unlabeled data fails to close the performance gap due to noisy proxy signals. 
In this paper, we propose learning robust real-world motion priors entirely from scalable simulation. 
We introduce SynFlow, a data generation pipeline for large-scale synthetic LiDAR scene flow.
Unlike prior works that prioritize sensor-specific realism, SynFlow employs a motion-oriented strategy to synthesize diverse kinematic patterns across 4,000 sequences ($\sim$940k frames), termed SynFlow-4k. This represents a $34\times$ scale-up in annotated volume over existing real-world benchmarks. 
Our experiments demonstrate that SynFlow-4k provides a highly domain-invariant motion prior. 
In a zero-shot regime, models trained only on our synthetic data generalize across multiple real-world benchmarks, comparable to in-domain supervised baselines on nuScenes and outperforming state-of-the-art methods on TruckScenes by $31.8\%$. 
Furthermore, SynFlow-4k serves as a label-efficient foundation: fine-tuning with only $5\%$ of real-world labels surpasses models trained from scratch on the full available budget. 
We open-source the pipeline and dataset to facilitate research in generalizable 3D motion estimation. More detail can be found at \url{https://kin-zhang.github.io/SynFlow}.

\end{abstract}

\section{Introduction}
\label{sec:intro}
Autonomous driving systems need to understand motion, not only recognize objects. 
Most 3D perception pipelines approach this through detection and tracking, where motion is inferred from objects belonging to predefined semantic categories~\cite{wang2023technical,pang2022simpletrack}. This object-centric formulation is effective for common traffic participants, but becomes less reliable for rare, ambiguous, or previously unseen movers. Scene flow offers a category-free alternative: it estimates dense, point-wise 3D motion directly from geometry, providing a motion-centric representation for downstream planning and interaction.

\begin{figure*}[t]
\centering
\includegraphics[trim=10 460 0 5, clip, width=\linewidth]{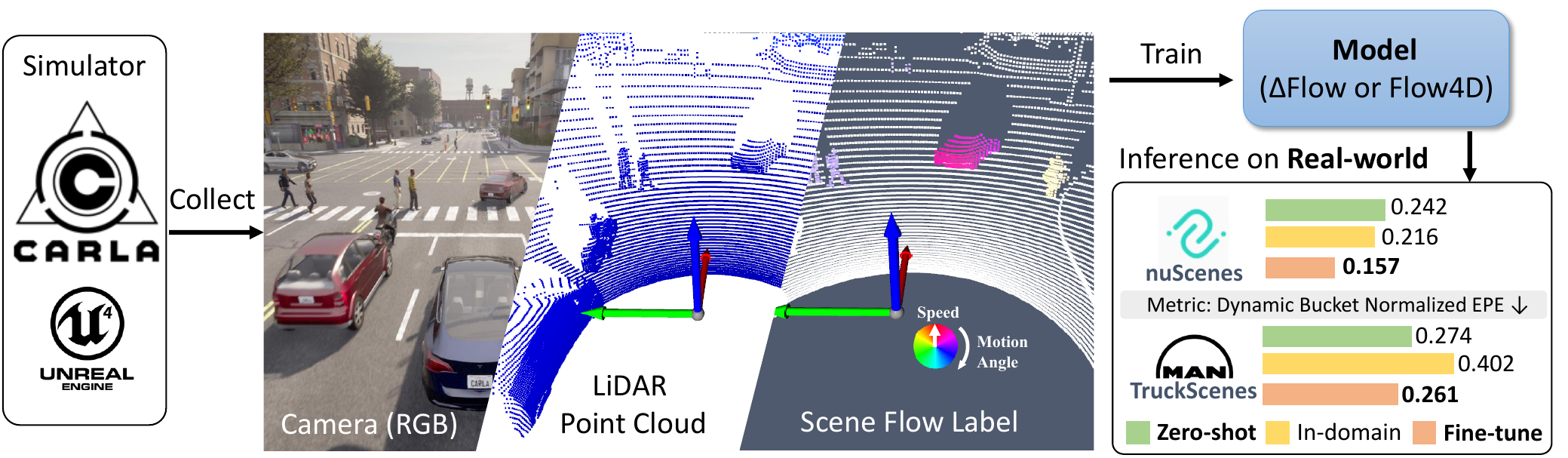}
\caption{Scaling up LiDAR Scene Flow with Synthetic Data. We present SynFlow, a data generation pipeline leveraging the CARLA simulator to synthesize diverse, perfectly labeled LiDAR scene flow data (center). 
While real-world datasets are often constrained by high annotation costs and limited scenario diversity, SynFlow provides a scalable source of dense, noise-free supervision for learning robust motion priors.
As shown in the results (right), models trained on SynFlow dataset achieve strong zero-shot generalization on real-world benchmarks and significantly outperform in-domain baselines when fine-tuned on a small subset of real data.}
\label{fig:cover}
\end{figure*}

Despite its importance, progress in LiDAR scene flow is constrained by the scarcity of reliable supervision~\cite{Argoverse2_2021}. 
Acquiring dense and accurate 3D motion annotations for real-world LiDAR data is expensive and practically infeasible at scale~\cite{zeroflow}. 
To mitigate this issue, recent works~\cite{yang2023vidar,zhang2024seflow,lin2025voteflow} have turned to self-supervised learning on large collections of unlabeled driving data. However, these approaches rely on geometric consistency assumptions, such as rigidity cluster or temporal alignment, to construct proxy self-supervision signals. These signals are inherently noisy and under-constrained, particularly in the presence of sensor sparsity, measurement noise. Consequently, scaling up unlabeled real-world data yields diminishing returns~\cite{zhang2024seflow}, leaving a substantial performance gap compared to fully supervised methods. 

This bottleneck motivates us to rethink how we source motion supervision.
Instead of relying on expensive human labels or noisy self-supervision, we explore whether robust motion priors can be learned entirely from scalable simulation.
We hypothesize that LiDAR scene flow learning depends primarily on capturing diverse kinematic physics rather than specific visual textures. Since simulators inherently generate precise rigid-body motion, they can provide reliable supervision even without perfect photorealism. Simulation further offers a unique advantage regarding data composition. Unlike real-world data collection~\cite{alibeigi2023zenseact,once21,sun2020scalability}, which is passive and limited to capturing events as they occur, simulation allows us to actively control the environment. We can easily vary sensor configurations, spawn diverse dynamic agents, and initialize traffic in specific scenarios. This ensures the model learns from a comprehensive range of motion patterns across various scenarios.
However, despite these advantages, existing synthetic LiDAR pipelines~\cite{yangrealistic,zhang2024resimad} have been optimized primarily for semantic realism or sensor-specific noise to support detection and segmentation tasks. 
To the best of our knowledge, no framework has yet been specifically designed to prioritize the dense kinematic complexity required for scene flow.

In this work, we introduce SynFlow, a data generation pipeline for LiDAR scene flow built upon the CARLA simulator~\cite{dosovitskiy2017carla} (\cref{fig:cover}). SynFlow adopts a motion-oriented generation strategy: rather than attempting to perfectly reproduce real-world sensor noise, we actively proceduralize traffic densities, aggressive speed regimes, and complex topological interactions across nine maps. 
Leveraging this pipeline, we release SynFlow-4k, a massive synthetic dataset comprising 4,000 fully annotated sequences ($\sim$940k frames). 
This is a 34$\times$ scale-up over the annotated frames of nuScenes~\cite{nuscenes} and 46$\times$ those of TruckScenes~\cite{fent2024man}, covering diverse road topologies, including roundabouts, intersections, and highways.

Through extensive experiments, we demonstrate that the model trained on SynFlow-4k generalizes zero-shot to different real-world sensors, comparable to in-domain supervised performance on nuScenes and beating the best in-domain results on TruckScenes by 31.8\%. As a pre-training foundation, fine-tuning on just 5\% of real labels already exceeds the performance of baselines trained from scratch on a budget four times larger.
Our dataset is available at \color{blue}\url{https://huggingface.co/datasets/KTH/SynFlow}\color{black}. 
Our primary contributions are as follows:

\begin{itemize}
    \item We propose \textbf{SynFlow}, the first synthesis pipeline designed specifically for LiDAR scene flow. It shifts the focus from sensor-specific realism to geometric and temporal interaction complexity to address the lack of dense motion supervision.
    
    \item We introduce \textbf{SynFlow-4k}, a large-scale synthetic dataset comprising 4,000 sequences ($\sim$940k frames) with dense, noise-free scene flow labels—representing a $34\times$ scale-up over existing real-world annotated resources.
    
    \item We demonstrate that \textit{SynFlow-4k} provides a robust, transferable motion prior: (1) models trained on it achieve strong zero-shot generalization across diverse real-world sensors; (2) it serves as a label-efficient pre-training foundation, significantly reducing real-world annotation demand; and (3) it complements real-world data by covering long-tail interactions that are rare in practice.
\end{itemize}

\section{Related Work}
\textbf{Synthetic Data as Scalable 3D Supervision.}
Synthetic data has increasingly been explored as a primary source of supervision for 3D learning~\cite{flythings3d,xie2024lrm,vggt,ren2025gen3c,vanhoorick2024gcd,wang2026vggtomega}. Early image-based scene flow datasets such as FlyingThings3D~\cite{flythings3d} demonstrated that dense motion fields can be learned from fully rendered 3D geometry. More recently, works such as LRM-Zero~\cite{xie2024lrm} and MegaSynth~\cite{jiang2025megasynth} show that large-scale synthetic data alone can learn transferable geometric priors and exhibit clear scaling behavior. 
These results suggest photorealistic real-world data may not be necessary for robust 3D representation learning, and suggest that controllable synthetic environments can serve as a scalable supervision source.

\noindent\textbf{Simulation in Autonomous Driving.}
In autonomous driving, CARLA~\cite{dosovitskiy2017carla} have been widely used to generate labeled data for perception tasks~\cite{carlasc22,cai2023analyzing,jiang2023optimizing} including detection, segmentation, and domain adaptation. 
Datasets such as SHIFT~\cite{sun2022shift} and CarlaScenes~\cite{kloukiniotis2022carlascenes} emphasize environmental diversity and cross-domain robustness, while reconstruction-simulation paradigms (e.g., ReSimAD~\cite{zhang2024resimad}) aim to bridge sensor gaps across domains. 
These efforts highlight the controllability and scalability of simulation for static perception and domain transfer. 
However, they focus on semantic realism or sensor alignment and rarely explore the kinematic complexity required for robust, zero-shot motion transfer to real-world environments.

\noindent\textbf{LiDAR Scene Flow and the Supervision Bottleneck.}
LiDAR scene flow~\cite{zhang2024gmsf,liu2024difflow3d,lin2025voteflow,lin2024icp,mambaflow} estimates point-wise 3D motion between consecutive point clouds and serves as a geometry-centric representation of dynamic environments. Due to the difficulty of obtaining dense real-world motion labels, most existing methods rely heavily on self-supervised objectives derived from geometric consistency~\cite{zhang2026teflow}, cycle constraints~\cite{vedder2024neural}, or rigidity assumptions~\cite{zhang2024seflow,hoffmann2025floxels}. While effective, such proxy self-supervision is inherently under-constrained and sensitive to occlusion, sparsity, and non-rigid motion. Scaling unlabeled real-world data alone does not fully close the gap to fully supervised performance.

While image-based synthetic datasets provide dense pixel-level motion annotations, prior efforts largely emphasize visual realism or domain alignment rather than motion supervision itself. We argue that 3D motion-centric tasks exhibit a different sim-to-real behavior compared to appearance-dominated tasks: since scene flow learning primarily depends on physically consistent object kinematics rather than texture or semantics, synthetic environments with accurate rigid-body states can provide transferable supervision signals for real-world LiDAR. Consequently, the key question is not whether simulation matches visual appearance, but whether scalable synthetic motion can serve as a reliable supervision source for LiDAR scene flow. Our work addresses this question by introducing a motion-oriented synthetic LiDAR data engine and systematically analyzing synthetic scaling and real-world fine-tuning behavior.

\section{Task and Preliminary}
\label{sec:task}
\paragraph{\textbf{Scene Flow Definition.}}
LiDAR scene flow aims to estimate the dense 3D motion field between consecutive point clouds in dynamic environments. Given two sequential scans, a source $\mathcal{P}_t = \{\mathbf{p}_i\}_{i=1}^{N_t} \subset \mathbb{R}^3$ and a target $\mathcal{P}_{t+1}$, the objective is to predict a per-point displacement field $\mathcal{F}_t = \{\mathbf{f}_i\}_{i=1}^{N_t}$. Each vector $\mathbf{f}_i \in \mathbb{R}^3$ represents the 3D translation of $\mathbf{p}_i$ from time $t$ to $t+1$ in continuous space.

\paragraph{\textbf{Temporal Context.}}
While the primary target is the forward flow from $\mathcal{P}_t$ to $\mathcal{P}_{t+1}$, modern estimators often leverage a temporal window of $h$ past frames $\{\mathcal{P}_{t-h}, \dots, \mathcal{P}_{t}, \mathcal{P}_{t+1}\}$ to improve motion reasoning. We denote a feed-forward scene flow estimator as $\Phi_\theta$ to learn the mapping:
\begin{equation}
\Phi_{\theta}: \{ \mathbf{T}_{\text{ego}}^{t-h \rightarrow t+1}\mathcal{P}_{t-h}, \dots, \mathbf{T}_{\text{ego}}^{t \rightarrow t+1}\mathcal{P}_{t}, \mathcal{P}_{t+1} \} \rightarrow \mathcal{F}_{t},
\end{equation}
where $\mathbf{T}_{\text{ego}}^{t' \rightarrow t+1} \in \mathbb{R}^{4 \times 4}$ is the odometry transformation matrix, aligning all historical point clouds into the coordinate frame of the target scan $\mathcal{P}_{t+1}$.

\paragraph{\textbf{Backbone Architecture.}}
In our experiments, we instantiate $\Phi_\theta$ with the $\Delta$Flow backbone~\cite{zhang2025deltaflow} as default. Following multi-frame designs, the backbone voxelizes each scan into sparse 3D features, aggregates temporal context into a compact representation, and applies a 3D sparse convolutional network (e.g., MinkUNet~\cite{choy20194d}) to extract motion-aware features. Finally, voxel features are interpolated back to points in $\mathcal{P}_t$ and decoded into the forward scene flow $\mathcal{F}_t$.

\begin{figure*}[t]
\centering
\includegraphics[trim=5 50 25 150, clip, width=\linewidth]{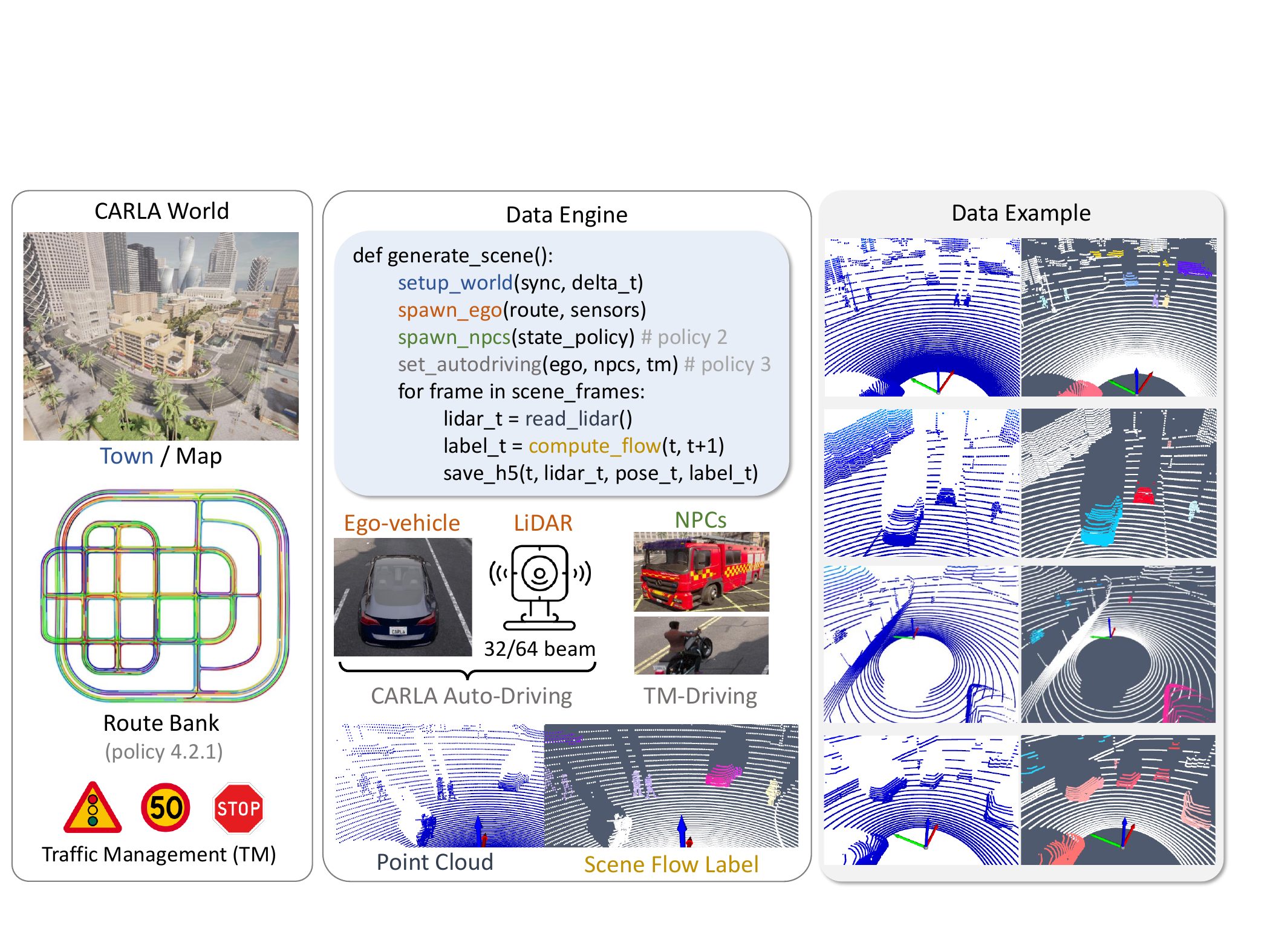}
\caption{Overview of our SynFlow pipeline and dataset examples. 
Left: A CARLA world provides diverse road topologies; we construct a route bank using topology-aware coverage to ensure broad spatial exploration and execute rollouts under Traffic Management (TM). 
Middle: our procedural data engine instantiates an ego vehicle with configurable LiDAR, spawns surrounding agents with controllable policies, and runs synchronized simulation steps; 
Right: Representative frames showing raw LiDAR point clouds alongside corresponding dense scene flow labels, spanning diverse traffic interactions and motion regimes.}
\label{fig:framework}
\end{figure*}

\section{Synthesizing the SynFlow Dataset}
\label{sec:synthesizing_xflow}
\subsection{Overview}
We introduce \textbf{SynFlow}, a synthesis pipeline designed to generate large-scale, kinematically diverse LiDAR datasets. 
Our design is \textit{motion-oriented}: rather than targeting visual photorealism, we prioritize the geometric and temporal complexity of multi-agent interactions. The generation process is guided by three core policies: (1) Topological Discretization Policy: To ensure the model learns diverse road geometries (e.g., roundabouts vs. highways); 
(2) Speed-Regime Coverage Policy: To broaden the displacement magnitude support by including highway-structured towns/routes and leveraging road-type-dependent speed limits under Traffic Manager control;
(3) Multi-Agent Interaction Policy: To enrich relative motion patterns by varying traffic density and agent behaviors, encouraging interaction-heavy scenarios beyond near-linear motion.
In this work, we instantiate these policies using the CARLA simulator~\cite{dosovitskiy2017carla}, which provides the high-fidelity physics and sensor models required to execute our kinematic rollouts. 

\subsection{Data Generation Pipeline}
\label{sec:xflow_policies}

The generation of SynFlow is an automated, iterative process. We initiate the pipeline by loading a target town topology and a predefined sensor configuration. For each town, the generation follows a sequence of ``Sampling $\rightarrow$ Rollout $\rightarrow$ Export'', governed by our three core policies.

\paragraph{\textbf{1. Topological Discretization Policy (Search Space Initialization).}}
The pipeline begins by discretizing the town's drivable topology into fine-grained lane segments \\
$(\texttt{road\_id}, \texttt{section\_id}, \texttt{lane\_id})$. 
We then initialize a \textit{route bank} $\mathcal{C}$ using a greedy search. A candidate route $\mathcal{R}$ is accepted only if it covers previously unvisited segments $|\mathcal{R}\setminus \mathcal{C}| > \tau$. This policy defines the ``where'' of our synthesis, ensuring geometric diversity by forcing the simulator to utilize long-tail road structures like Town04's highway loops and Town06's junctions.

\paragraph{\textbf{2. Speed-Regime Coverage Policy (State Initialization).}}
After selecting a route, we initialize the scene by spawning the ego vehicle and surrounding agents (NPCs) on valid waypoints along the route neighborhood.
Rather than hand-tuning a target speed, we leverage the road-type-dependent speed limits and CARLA Traffic Manager control to induce diverse velocity regimes.
Importantly, we include towns/routes containing highway-like segments (e.g., loop highways and long multi-lane roads), which naturally produce high-speed motion tails,
complementing low-speed stop-and-go urban traffic.
This policy defines the ``how fast'' of our synthesis and improves coverage of large displacements without requiring expensive per-scene parameter search.

\paragraph{\textbf{3. Multi-Agent Interaction Policy (Dynamic Execution).}}
During rollouts, we induce interaction-heavy scenes by varying local traffic density and agent behaviors under CARLA Traffic Manager control.
This encourages diverse interaction regimes (e.g., merges, overtakes, and braking events) and yields complex non-linear relative motions
that are essential for learning robust scene flow.

\paragraph{\textbf{Simulation rollout and export.}}
Given the route, kinematic states, and interaction regimes sampled by the policies above, we execute data collection in CARLA.
For each town and LiDAR configuration, we initialize CARLA in a deterministic synchronous mode with a fixed simulation step $\Delta t=0.1$, spawn the ego vehicle at the sampled route start, and attach a 32/64-beam LiDAR.
We populate the scene with heterogeneous NPCs (different vehicles and pedestrians) within a local neighborhood and control them via CARLA Traffic Manager.
To avoid low-motion segments caused by long stops or traffic deadlocks, we apply a lightweight deadlock-resolution rule: if the ego remains stationary beyond a short threshold due to a blocked intersection, we override the local traffic signal to release the blockage and resume the rollout, thereby maintaining a high ratio of dynamic frames.
At every step, we compute ground truth scene flow supervision (Sec.~\ref{sec:xflow_oracle}) and store the rollout in HDF5 containers~\cite{The_HDF_Group_Hierarchical_Data_Format}. 
Each timestamp entry contains the LiDAR point cloud $\mathcal{P}_t$, ego pose, aligned flow labels $\mathcal{F}_t$, validity masks, and per-point instance metadata for training and evaluation.

\subsection{Scene Flow Label Generation}
\label{sec:xflow_oracle}
After generating the dynamic scenarios, we leverage the simulator’s ground-truth physical states to derive noiseless, per-point scene flow labels for all dynamic objects in the scene. 

\paragraph{\textbf{Simulator-provided information.}} 
At every timestep $t$, the simulator exposes the full world-coordinate rigid-body pose $\mathbf{T}^t_k \in SE(3)$ for each tracked agent $k$, including vehicles, pedestrians, and cyclists. 
In addition, the simulator provides each raw point $\mathbf{p}_i \in \mathcal{P}_t$ with a per-point instance identifier $u_i$ that is consistent across timesteps. 
This allows us to associate every LiDAR point with a specific physical object and derive where that point moves between $t$ and $t+1$.

\paragraph{\textbf{Point-to-agent tag assignment.}}
Since the per-point instance tag $u_i$ is not aligned with the simulator actor ID $k$, we resolve the correspondence by majority voting over tags within the bounding box of agent $k$ at time $t$:
\begin{equation}
\hat{u}^t_k=\operatorname*{argmax}_{u}\sum_{i:\,\mathbf{p}_i \in \operatorname{bbox}(k,t)}\mathbf{1}[u_i=u],
\qquad
\mathcal{M}^t_k=\{\, \mathbf{p}_i \in \mathcal{P}_t \mid u_i=\hat{u}^t_k \,\}.
\end{equation}
where $\mathcal{M}^t_k$ is the set of LiDAR points in $\mathcal{P}_t$ that belong to agent $k$.

\paragraph{\textbf{Rigid-body flow derivation.}} For each agent $k$, the simulator provides its world-coordinate pose at both timesteps, $\mathbf{T}^t_k$ and $\mathbf{T}^{t+1}_k$. 
The LiDAR scan $\mathcal{P}_t$ gives us the observed point positions at time $t$, but the corresponding positions at $t+1$ are never directly observed as they must be inferred. For any point $\mathbf{p}_i \in \mathcal{M}_k^t$, we exploit the known rigid-body motion of agent $k$ to estimate where that surface element moves:
\begin{equation}
\mathbf{p}^{\star}_i = \mathbf{T}^{t+1}_k (\mathbf{T}^{t}_k)^{-1} \mathbf{p}_i.
\end{equation}
Here $(\mathbf{T}^{t}_k)^{-1}$ maps $\mathbf{p}_i$ into the agent's local body frame, and $\mathbf{T}^{t+1}_k$ places it back into world space at the agent's pose at $t+1$. The ground truth flow vector is then $\mathbf{f}_i = \mathbf{p}^{\star}_i - \mathbf{p}_i$.

\section{Training Scene Flow Estimation}
\label{sec:training}
In this section, we discuss the utilization of our synthesized SynFlow dataset to train a flow estimation model and define the protocols for evaluating its quality. We first describe the preparation and scaling of the SynFlow dataset, followed by our two-stage training strategy for real-world adaptation and the technical implementation details.

\subsection{SynFlow Dataset Preparation}
\label{sec:dataset_prep}

Following the procedure described in Sec.~\ref{sec:xflow_policies}, we generate the SynFlow dataset consisting of 4k fully annotated LiDAR sequences (SynFlow-4k), totaling 939,083 frames. 
Data collection is performed on a desktop system equipped with an Intel i7-12700KF processor and a NVIDIA RTX 3090 GPU. Generating one sequence takes 3–6 minutes depending on the scene configuration.
To study how synthetic supervision scales, we construct four training splits (1k, 2k, 3k, and 4k sequences) while controlling for map and sensor factors.

\begin{table}[h]
\centering
\small
\setlength{\tabcolsep}{6pt}
\caption{Summary of SynFlow-4k data splits. Scaling annotated volume via rollouts across CARLA towns and route banks. Training sets comprise mixed 32/64-beam LiDAR sequences.}
\begin{tabular}{c|l|l|l}
\toprule
\textbf{Split} & \textbf{\#Annotated Frames} & \textbf{Composition} & \textbf{Scenarios} \\
\midrule
1k & 271{,}148 & Town06, 07, 10 &
arterials, complex junctions, rural roads \\
2k & 449{,}407 & Town01--05 &
roundabouts, multi-lane intersections \\
3k & 720{,}555 & 1k + 2k &
(1k + 2k) \\
4k & 939{,}083 & 2k + Town12 &
mixed zones (urban/suburban/rural) \\
\bottomrule
\end{tabular}
\label{tab:scaling_splits}
\end{table}

Each split aggregates complete rollouts across multiple towns and route banks, ensuring broad topological coverage. To maintain balanced map composition across scales, we subsample routes from Town12 (the largest route bank) and include them only in the 4k split. All splits utilize a mixture of 32/64-beam LiDAR configurations to ensure sensor-agnostic motion learning. The exact composition and dominant scene characteristics of these splits are summarized in Tab.~\ref{tab:scaling_splits}. 
Sample SynFlow-4k visualizations are in Sec~\ref{sec:qualitative_com}, with more detailed video samples at \url{https://kin-zhang.github.io/SynFlow}.

\subsection{Real-World Evaluation Strategy}
\label{sec:real_finetune}
To evaluate the quality of SynFlow-4k and its utility for real-world tasks, we define two primary evaluation regimes:
\textit{1) Zero-shot Generalization:} 
We evaluate models trained only on SynFlow-4k to measure how well they transfer to different real-world datasets without any domain adaptation;
\textit{2) Label-Efficient Fine-tuning:} To quantify the reduction in required real-world annotations, we use SynFlow-4k as a pre-training initialization. Models are first trained on our synthetic ground truth labels and then fine-tuned on restricted subsets (5--20\%) of the target real-world benchmarks. By starting from the SynFlow-4k checkpoint and continuing supervised training only on limited real samples, we evaluate the model's ability to transfer a mature motion prior to data-scarce real-world environments.

\subsection{Implementation Details}
\label{sec:implementation}
\noindent\textbf{Backbone and Loss.} We adopt the $\Delta$Flow~\cite{zhang2025deltaflow} backbone and follow its default architecture settings, using 5-frame LiDAR sequences voxelized at 0.15m within a 38.4m range. We train with the supervised scene flow loss introduced in $\Delta$Flow: motion-awareness, category-balanced, and instance-consistency terms. For synthetic data, ground-truth is obtained via simulator states with the target construction described in Sec.~\ref{sec:xflow_oracle}.

\noindent\textbf{Optimization and Augmentation.} 
We train for 15 epochs using Adam with a learning
rate of 0.002 and a total batch size of 20. 
We apply standard augmentations (e.g., z-height perturbation, random x-y flips) without additional domain adaptation.

\noindent\textbf{Evaluation Datasets.} We evaluate our method on three real-world LiDAR benchmarks. \textit{nuScenes}~\cite{nuscenes} provides urban driving scenes captured with a 32-beam LiDAR; its training split contains 700 scenes totaling 137,575 frames. \textit{TruckScenes}~\cite{fent2024man} focuses on long-haul highway driving with a dual 64-beam LiDAR platform; its training split contains 524 scenes and 101,902 frames. We choose nuScenes and TruckScenes as our primary benchmarks for label efficiency because their annotated ground-truth is available for only a small subset ($\sim$20\%) of the full sequences. 
Finally, the \textit{Aeva}~\cite{aevascenes} dataset provides FMCW LiDAR sequences covering both urban and highway driving; we evaluate on 67 sequences after performing a flow generation consistency check.

\noindent\textbf{Evaluation Metrics.} 
We follow prior work and report \textit{Dynamic Bucket-Normalized EPE}~\cite{khatri2024can} as our primary metric, which normalizes errors by motion magnitude across velocity buckets. We additionally report \textit{Three-way EPE}~\cite{chodosh2023re} for completeness to facilitate comparison with existing methods.

\noindent\textbf{Baselines.} 
We compare against representative LiDAR scene flow baselines from prior works, including both self-supervised methods (SeFlow~\cite{zhang2024seflow}, VoteFlow~\cite{lin2025voteflow}, SeFlow++~\cite{zhang2025himo}, TeFlow~\cite{zhang2026teflow}) and fully supervised feed-forward estimators (DeFlow~\cite{zhang2024deflow}, Flow4D~\cite{kim2024flow4d}, $\Delta$Flow~\cite{zhang2025deltaflow}). All methods are trained with the best configurations as reported in their papers in the in-domain real-world datasets.

\begin{table*}[t]
\setlength{\tabcolsep}{0.5em}
\centering
\caption{
Performance on the \textbf{nuScenes} validation set.
\textit{Supervision}: \textit{unlab.}~=~self-supervised on unlabeled real sequences;
\textit{lab.}~=~supervised on available (20\%) labeled real data;
\textit{synth.}~=~supervised on SynFlow synthetic data only (no real data).
\textbf{Bold} denotes the best result in each column. 
}
\label{tab:nuscenes}
\resizebox{\linewidth}{!}{
\begin{tabular}{lc|ccccc|rrrr}
\toprule
\multirow{2}{*}{Methods}
  & \multirow{2}{*}{Supervision}
  & \multicolumn{5}{c|}{Dynamic Bucket-Normalized $\downarrow$}
  & \multicolumn{4}{c}{\small Three-way EPE (cm) $\downarrow$} \\
  & & Mean & CAR & OTHER & PED. & VRU
    & \small Mean & \small FD & \small FS & \small BS \\
\midrule
Ego Motion Flow & -- & 1.000 & 1.000 & 1.000 & 1.000 & 1.000
                     & \small 12.34 & \small 35.94 & \small 1.07 & \small 0.00 \\
\midrule
\multicolumn{11}{l}{\textit{\footnotesize Self-supervised (100\% unlabeled real-world data)}} \\[1pt]
SeFlow~\cite{zhang2024seflow}         & 100\% unlab.
  & 0.544 & 0.396 & 0.635 & 0.726 & 0.419
  & \small 8.19 & \small 16.15 & \small 3.97 & \small 4.45 \\
VoteFlow~\cite{lin2025voteflow}       & 100\% unlab.
  & 0.538 & 0.355 & 0.605 & 0.780 & 0.410
  & \small 7.80 & \small 15.65 & \small 3.51 & \small 4.24 \\
SeFlow++~\cite{zhang2025himo}         & 100\% unlab.
  & 0.509 & 0.327 & 0.583 & 0.716 & 0.409
  & \small 6.13 & \small 14.59 & \small 1.96 & \small 1.86 \\
TeFlow~\cite{zhang2026teflow}         & 100\% unlab.
  & 0.395 & 0.303 & 0.461 & 0.474 & 0.344
  & \small 4.64 & \small 10.92 & \small 1.49 & \small 1.51 \\
\midrule
\multicolumn{11}{l}{\textit{\footnotesize Fully supervised (20\% labeled real-world data)}} \\[1pt]
DeFlow~\cite{zhang2024deflow}         & 20\% lab.
  & 0.314 & 0.163 & 0.286 & 0.533 & 0.275
  & \small 3.98 & \small 6.99 & \small 3.45 & \small 1.50 \\
Flow4D~\cite{kim2024flow4d}           & 20\% lab.
  & 0.279 & 0.204 & 0.312 & 0.379 & 0.222
  & \small 3.82 & \small 8.05 & \small 1.82 & \small 1.58 \\
$\Delta$Flow~\cite{zhang2025deltaflow}& 20\% lab.
  & 0.216 & 0.138 & 0.219 & 0.327 & 0.181
  & \small 2.33 & \small 4.83 & \small 1.37 & \small 0.79 \\
\midrule
\multicolumn{11}{l}{\textit{\footnotesize SynFlow pre-training (ours, synthetic data only $\rightarrow$ optional fine-tune)}} \\[1pt]
SynFlow-4k          & 0\% (synth. only)
  & 0.242 & 0.177 & 0.307 & 0.300 & 0.183
  & \small 2.60 & \small 6.06 & \small 1.28 & \small 0.46 \\
SynFlow-4k + FT     & 20\% lab.
  & \textbf{0.157} & \textbf{0.110} & \textbf{0.173} & \textbf{0.247} & \textbf{0.098}
  & \small\textbf{1.65} & \small\textbf{3.48} & \small\textbf{1.19} & \small\textbf{0.29} \\
\bottomrule
\end{tabular}
}
\end{table*}

\begin{table*}[t]
\setlength{\tabcolsep}{0.5em}
\centering
\caption{
Performance on the \textbf{TruckScenes} validation set.
TruckScenes features large commercial vehicles with distinct LiDAR configurations unseen during SynFlow pre-training, providing a stringent test of cross-domain zero-shot generalization.
Notation follows Tab.~\ref{tab:nuscenes}.
}
\label{tab:truckscenes}
\resizebox{\linewidth}{!}{
\begin{tabular}{lc|ccccc|rrrr}
\toprule
\multirow{2}{*}{Methods}
  & \multirow{2}{*}{Supervision}
  & \multicolumn{5}{c|}{Dynamic Bucket-Normalized $\downarrow$}
  & \multicolumn{4}{c}{\small Three-way EPE (cm) $\downarrow$} \\
  & & Mean & CAR & OTHER & PED. & VRU
    & \small Mean & \small FD & \small FS & \small BS \\
\midrule
Ego Motion Flow & -- & 1.000 & 1.000 & 1.000 & 1.000 & 1.000
                     & \small 61.43 & \small 184.30 & \small 0.00 & \small 0.00 \\
\midrule
\multicolumn{11}{l}{\textit{\footnotesize Self-supervised (100\% unlabeled real-world data)}} \\[1pt]
SeFlow~\cite{zhang2024seflow}         & 100\% unlab.
  & 0.681 & 0.494 & 0.752 & 0.886 & 0.591
  & \small 37.41 & \small 103.97 & \small 1.56 & \small 6.68 \\
VoteFlow~\cite{lin2025voteflow}       & 100\% unlab.
  & 0.680 & 0.517 & 0.737 & 0.895 & 0.573
  & \small 36.47 & \small 103.48 & \small 1.29 & \small 4.64 \\
SeFlow++~\cite{zhang2025himo}         & 100\% unlab.
  & 0.653 & 0.519 & 0.738 & 0.860 & 0.497
  & \small 33.35 & \small 95.62  & \small 1.20 & \small 3.23 \\
TeFlow~\cite{zhang2026teflow}         & 100\% unlab.
  & 0.425 & 0.254 & 0.455 & 0.636 & 0.353
  & \small 41.97 & \small 116.55 & \small 1.32 & \small 8.06 \\
\midrule
\multicolumn{11}{l}{\textit{\footnotesize Fully supervised (20\% labeled real-world data)}} \\[1pt]
DeFlow~\cite{zhang2024deflow}         & 20\% lab.
  & 0.570 & 0.180 & 0.410 & 0.970 & 0.730
  & \small 7.30  & \small 16.47 & \small 1.67 & \small 3.77 \\
Flow4D~\cite{kim2024flow4d}           & 20\% lab.
  & 0.456 & 0.176 & 0.351 & 0.885 & 0.413
  & \small 16.14 & \small 44.87 & \small 1.71 & \small 1.85 \\
$\Delta$Flow~\cite{zhang2025deltaflow}& 20\% lab.
  & 0.402 & 0.196 & 0.400 & 0.690 & 0.323
  & \small 7.28  & \small 16.26 & \small 1.36 & \small 4.52 \\
\midrule
\multicolumn{11}{l}{\textit{\footnotesize SynFlow pre-training (ours, synthetic data only $\rightarrow$ optional fine-tune)}} \\[1pt]
SynFlow-4k          & 0\% (synth. only)
  & 0.274 & 0.109 & 0.220 & 0.467 & 0.300
  & \small 25.25 & \small 74.16 & \small 1.02 & \small 0.58 \\
SynFlow-4k + FT     & 20\% lab.
  & \textbf{0.266} & \textbf{0.082} & \textbf{0.232} & \textbf{0.464} & \textbf{0.285}
  & \small\textbf{6.75} & \small\textbf{15.70} & \small\textbf{0.98} & \small\textbf{3.58} \\
\bottomrule
\end{tabular}
}
\end{table*}

\section{Results}
\label{sec:results}
\subsection{Baseline Comparison}
\cref{tab:nuscenes} and~\cref{tab:truckscenes} report the primary results on nuScenes and TruckScenes across three learning regimes: self-supervised learning on the full unlabeled training split, supervised training on the human-labeled data, and zero-shot transfer from synthetic training.

\paragraph{\textbf{Zero-shot Transfer.}}
Without observing any real-world data, SynFlow-4k achieves strong zero-shot generalization on both benchmarks. On nuScenes, our zero-shot prior achieves a Dynamic EPE of 0.242, outperforming the best self-supervised baseline (TeFlow, 0.395) by 38.7\% and approaching the performance of supervised methods. On Truck-\\-Scenes, it attains 0.274, outperforming TeFlow (0.425) by 35.5\% and even surpassing the SOTA supervised $\Delta$Flow baseline (0.402) by 31.8\%.  
These results across different sensors suggest that physically consistent motion transfers well for LiDAR scene flow.
Our motion-oriented synthetic data effectively bridges the sim-to-real gap without explicit domain adaptation.

\paragraph{\textbf{Fine-tuning Transfer.}}
As shown in \cref{tab:ablation_finetune}, pre-training on SynFlow-4k yields a stronger motion prior, leading to more effective downstream fine-tuning.
Fine-tuning on 20\% of real labels achieves 0.157 on nuScenes, outperforming the supervised $\Delta$Flow baseline (0.216) by 27.3\%. On TruckScenes, the gain reaches 33.8\% (0.266 vs. 0.402). These gains are achieved in the practical regime where dense annotation is expensive and only a fraction of frames can be labeled.
Overall, the results indicate that SynFlow-4k provides a strong initialization that significantly reduces the demand for real-world labels, 
fine-tuning on just 5\% of real labels already exceeds baselines trained from scratch on $4\times$ more data.

\subsection{Ablation Studies}
\begin{figure}[t]
  \centering
  \includegraphics[width=\linewidth]{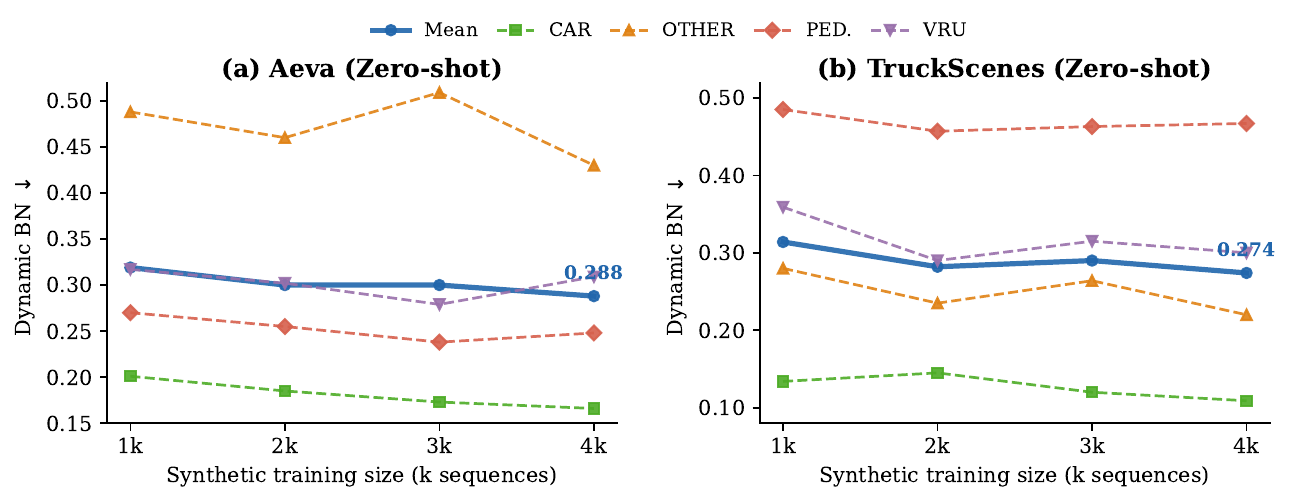}
  \caption{
    Zero-shot scaling performance (1k–4k sequences). Evaluation on Aeva (a) and TruckScenes (b) using Dynamic Bucket-Normalized EPE (lower is better).
    Solid blue line indicates the overall mean; dashed lines represent per-category breakdowns.
  }
  \label{fig:ablation_scale}
\end{figure}

\paragraph{\textbf{Zero-shot Scaling.}}
To understand the effect of synthetic data scale on zero-shot performance, we evaluate training volumes ranging from 1k to 4k sequences (Fig.~\ref{fig:ablation_scale}). Performance improves consistently across both benchmarks, with the most significant gains observed between 1k and 2k. Beyond this point, accuracy begins to stabilize, suggesting the core motion distribution for rigid-body kinematics is effectively captured within the 4k split. Among categories, \textit{CAR} benefits most from additional data.
Its rigid and predictable kinematics transfer well from simulation, reaching 0.109 on TruckScenes at 4k without any real labeled data. 
These results confirm that synthetic scale is a reliable lever for improving motion priors, particularly for vehicle-class objects.

\paragraph{\textbf{Fine-tuning Scaling.}}
To evaluate how real-world data scale affects adaptation, we ablate the labeling budget required for fine-tuning.
As shown in Tab.~\ref{tab:ablation_finetune}, SynFlow-4k provides a high-fidelity initialization that significantly reduces label demand. 
On nuScenes, fine-tuning our prior with just 5\% of real labels (0.201) already outperforms the supervised baseline trained from scratch on 20\% of the data (0.216). Increasing the fine-tuning budget to 10\% and 20\% further reduces Dynamic Bucketed-Normalized EPE to 0.175 and 0.157, respectively.
On TruckScenes, this advantage is even more pronounced: our zero-shot model already outperforms the 20\% supervised baseline (0.274 vs. 0.402), and adding a 10\% labeling budget further reduces error to 0.261. Across both datasets, the most rapid performance gains occur within the first 10\% of real-world exposure, with further improvements scaling more gradually up to 20\%. This demonstrates that SynFlow-4k serves as a strong pre-training foundation, significantly reducing the need for real-world labels while outperforming training from scratch.

\begin{table}[t]
\setlength{\tabcolsep}{0.8em}
\centering
\caption{
Fine-tuning scaling on nuScenes (upper) and TruckScenes (lower). Comparison between from-scratch supervised baselines and SynFlow-4k across varying real-world label budgets. Zero-shot (0\%) denotes purely synthetic supervision; +FT rows indicate fine-tuning on the specified real-world fraction. \textbf{Bold} denotes best per column.
}
\label{tab:ablation_finetune}
\resizebox{0.9\linewidth}{!}{
\begin{tabular}{lc|ccccc}
\toprule
Methods & Real labels
  & Mean & CAR & OTHER & PED. & VRU \\
\midrule
\multicolumn{7}{l}{\textit{\footnotesize nuScenes validation set}} \\[1pt]
$\Delta$Flow~\cite{zhang2025deltaflow} & 20\% & 0.216 & 0.138 & 0.219 & 0.327 & 0.181 \\
\cmidrule{1-7}
SynFlow-4k (Zero-shot)  & 0\%  & 0.242 & 0.177 & 0.307 & 0.300 & 0.183 \\
SynFlow-4k + FT         & 5\% & 0.201 & 0.151 & 0.240 & 0.272 & 0.139 \\
SynFlow-4k + FT         & 10\% & 0.175 & 0.127 & 0.201 & 0.253 & 0.119 \\
SynFlow-4k + FT         & 20\% & \textbf{0.157} & \textbf{0.110} & \textbf{0.173} & \textbf{0.247} & \textbf{0.098} \\
\midrule
\midrule
\multicolumn{7}{l}{\textit{\footnotesize TruckScenes validation set}} \\[1pt]
$\Delta$Flow~\cite{zhang2025deltaflow} & 20\% & 0.402 & 0.196 & 0.400 & 0.690 & 0.323 \\
\cmidrule{1-7}
SynFlow-4k (Zero-shot)  & 0\%  & 0.274 & 0.109 & 0.220 & 0.467 & 0.300 \\
SynFlow-4k + FT         & 5\% & 0.271 & 0.101 & 0.214 & 0.471 & 0.299 \\
SynFlow-4k + FT         & 10\% & \textbf{0.261} & 0.088 & \textbf{0.215} & 0.478 & \textbf{0.262} \\
SynFlow-4k + FT         & 20\% & 0.266 & \textbf{0.082} & 0.232 & \textbf{0.464} & 0.285 \\
\bottomrule
\end{tabular}
}
\end{table}

\begin{table}[t]
\setlength{\tabcolsep}{0.65em}
\centering
\caption{
Complementarity of SynFlow-4k and UniFlow. Zero-shot evaluation on the Aeva dataset. UniFlow~\cite{li2025uniflowzeroshotlidarscene} is trained on merged real-world data (Argo-v2~\cite{Argoverse2_2021}, nuScenes, Waymo). Combining synthetic and real-world sources consistently outperforms individual training.
}
\label{tab:analysis_uniflow}
\resizebox{0.95\linewidth}{!}{
\begin{tabular}{l|ccccc|rrrr}
\toprule
\multirow{2}{*}{Training data}
& \multicolumn{5}{c|}{Dynamic Bucket-Normalized $\downarrow$}
& \multicolumn{4}{c}{\small Three-way EPE (cm) $\downarrow$} \\
& Mean & CAR & OTHER & PED. & VRU
& \small Mean & \small FD & \small FS & \small BS \\
\midrule
UniFlow only  & 0.303 & 0.112 & \textbf{0.359} & 0.398 & 0.344 & \small 6.65 & \small 13.17 & \small 5.16 & \small 1.62 \\
SynFlow-4k only & 0.288 & 0.166 & 0.430 & \textbf{0.248} & 0.309 & \small 9.08 & \small 21.41 & \small 4.88 & \small 0.94 \\
Both          & \textbf{0.263} & \textbf{0.102} & 0.401 & 0.251 & \textbf{0.298} & \small\textbf{5.95} & \small\textbf{13.15} & \small\textbf{3.85} & \small\textbf{0.84} \\
\bottomrule
\end{tabular}
}
\end{table}

\paragraph{\textbf{Synthetic-Real Complementarity.}}
To investigate how synthetic supervision can further push the limits of zero-shot transfer, we evaluate SynFlow-4k in combination with UniFlow~\cite{li2025uniflowzeroshotlidarscene}: a state-of-the-art model pre-trained on a massive union of real-world datasets (Argo-v2, nuScenes, and Waymo). This experiment tests whether synthetic data is merely a ``low-cost substitute'' for real labels or an orthogonal source of knowledge (Tab.~\ref{tab:analysis_uniflow}). 
We find that combining both sources (0.263) consistently outperforms either individual model, yielding a striking 37\% improvement in the PED. category (0.398 $\to$ 0.251). 
This large gain suggests that even mega-scale real datasets lack the kinematic density needed for small, dynamic agents. While UniFlow captures real sensor characteristics (e.g., beam layouts, range-dependent sparsity, and measurement noise), SynFlow-4k contributes dense, oracle kinematics for ``long-tail'' interactions that are rarely seen in real-world logs. 
These results demonstrate that synthetic and real-world pre-training are highly complementary: simulation provides the fundamental motion ``rules'', while real data provides the sensor ``realism''.

\paragraph{\textbf{Synthetic Generation Strategy.}}
To investigate the architectural design of \textit{SynFlow} and identify the key factors driving its strong generalization, we isolate our data generation policies using a 1k-sequence ablation split (Tab.~\ref{tab:ablation_design}).
We find that our Topological Discretization Policy (Policy 1), represented by Route Coverage in the table, provides the largest individual gain (0.346 $\to$ 0.330). By replacing random route sampling with a greedy search over unvisited lane segments, the model encounters a more diverse set of road structures, leading to more robust motion contexts. 
Furthermore, including Highway towns to support our Speed Regime Policy (Policy 2) provides a complementary benefit (0.330 $\to$ 0.319). Its impact is most visible in the Three-way EPE metrics, where it significantly reduces EPE Foreground Dynamic (FD) error (50.21 $\to$ 48.17,cm). This confirms that exposing the model to the high-speed displacements found in highway loops is essential for generalizing to the high-velocity tails of real-world datasets. The combination of both policies ensures that SynFlow dataset covers both long-tail road structures and a broad support of motion magnitudes.

\begin{table}[t]
\setlength{\tabcolsep}{0.65em}
\centering
\caption{
  Ablation of data generation design choices (1k sequences trained), zero-shot evaluated on Aeva dataset.
  ``Topology Coverage (Policy~1)'' toggles greedy lane-segment coverage route bank; otherwise, random-start fixed-length routes are used.
  ``Speed Regime (Policy~2)'' toggles whether highway-structured towns (Town04, Town06, Town07) are included; otherwise only urban towns are used.
}
\label{tab:ablation_design}
\resizebox{0.95\linewidth}{!}{
\begin{tabular}{cc|ccccc|rrrr}
\toprule
\multirow{2}{*}{\begin{tabular}[c]{@{}c@{}}P1:\\Topology\end{tabular}}
& \multirow{2}{*}{\begin{tabular}[c]{@{}c@{}}P2:\\Speed\end{tabular}} 
& \multicolumn{5}{c|}{Dynamic Bucket-Normalized $\downarrow$}
& \multicolumn{4}{c}{\small Three-way EPE (cm) $\downarrow$} \\
& & Mean & CAR & OTHER & PED. & VRU
  & \small Mean & \small FD & \small FS & \small BS \\
\midrule
            &             & 0.346 & 0.225 & 0.512 & 0.311 & 0.334             & 18.96 & 52.34 & 3.51 & 1.02                \\ 
            & \checkmark  & 0.331 & 0.211 & 0.487 & 0.282 & 0.342             & 18.45 & 51.25 & 3.10 & 1.01  \\
\checkmark  &             & 0.330 & 0.215 & 0.501 & 0.281 & 0.324 & \small 17.89 & \small 50.21 & \small 2.82 & \small \textbf{0.65} \\
\checkmark  & \checkmark  & \textbf{0.319} & \textbf{0.201} & \textbf{0.488} & \textbf{0.270} & \textbf{0.317} & \small\textbf{17.11} & \small\textbf{48.17} & \small\textbf{2.52} & \small 0.66 \\
\bottomrule
\end{tabular}
}
\end{table}

\begin{table}[t!]
\centering
\small
\setlength{\tabcolsep}{0.65em}
\caption{Ablation of backbone architectures. Zero-shot evaluation on Aeva after training on SynFlow-4k dataset. The consistent improvement across different estimators demonstrates the backbone-agnostic transferability of our synthetic supervision.}
\label{tab:aeva_backbone_agnostic}
\resizebox{0.95\linewidth}{!}{
\begin{tabular}{l|ccccc|cccc}
\toprule
\multirow{2}{*}{Backbone} &
\multicolumn{5}{c|}{Dynamic Bucket-Normalized $\downarrow$} &
\multicolumn{4}{c}{Three-way EPE (cm) $\downarrow$} \\
& MEAN & CAR & OTHER & PED. & VRU & Mean & FD & FS & BS \\
\midrule
Ego Motion Flow & 1.000 & 1.000 & 1.000 & 1.000 & 1.000 & 44.31 & 132.00 & 0.92 & 0.00 \\
\midrule
\multicolumn{6}{l}{\textit{\footnotesize Trained on SynFlow-4k}} \\[1pt]
DeFlow          & 0.426 & 0.249 & 0.514 & 0.448 & 0.500 & 10.99 & 25.85  & 6.60 & 0.84 \\
Flow4D          & 0.308 & 0.173 & 0.452 & 0.286 & 0.320 & 9.37  & 21.87  & 6.00 & 0.94 \\
$\Delta$Flow       & 0.288 & 0.166 & 0.430 & 0.248 & 0.309 & 9.08   & 21.41  & 4.88 & 0.94 \\
\bottomrule
\end{tabular}}
\end{table}

\begin{figure}[h!]
\centering
\includegraphics[width=\linewidth]{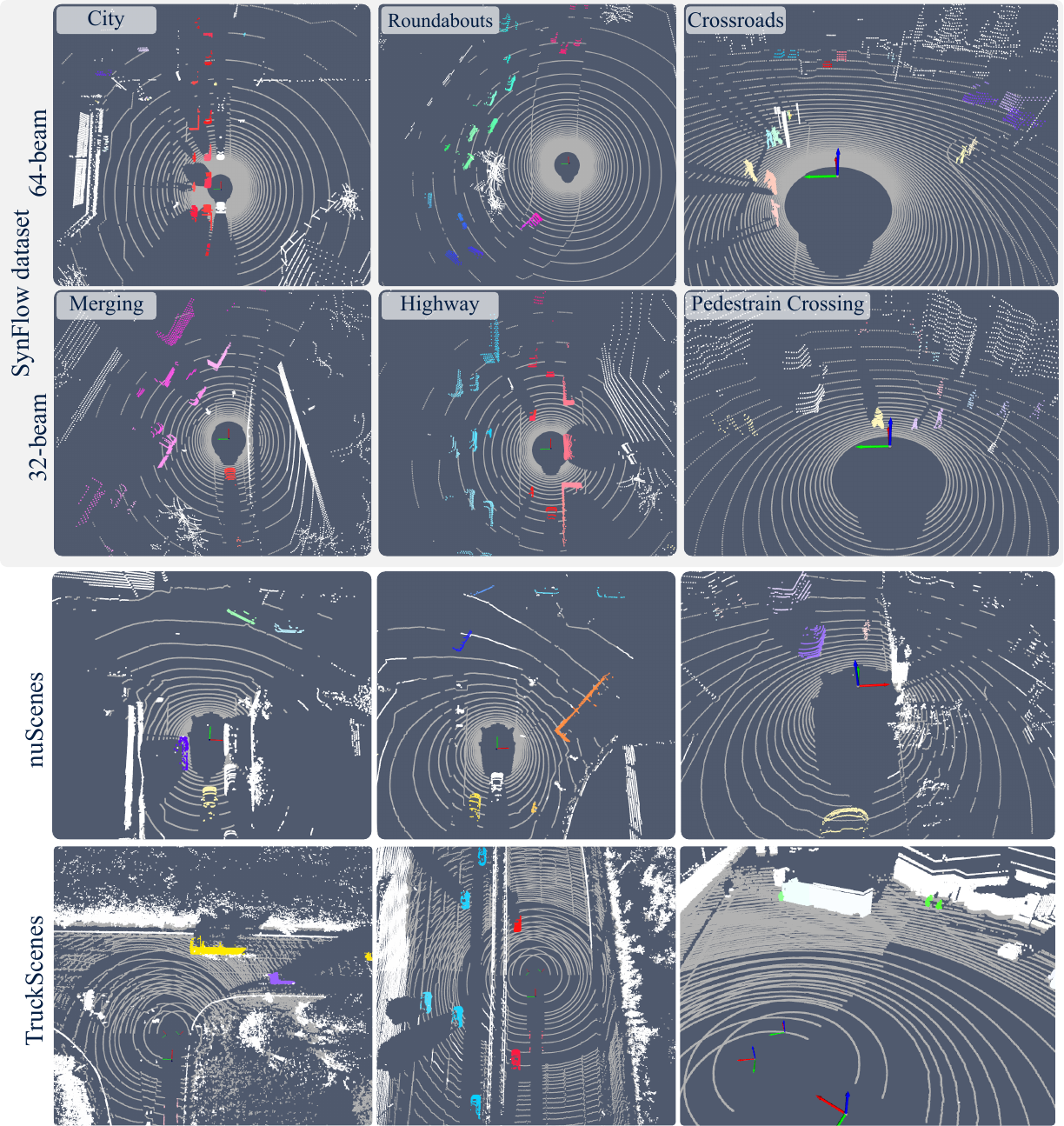}
\caption{\textbf{Dataset visualization.} Top: SynFlow-4k samples under 64-beam (row 1) and 32-beam (row 2) configurations, spanning city, roundabout, highway, and merging scenarios. 
Per-point scene flow labels are rendered as colored vectors. Direction is encoded as hue, and magnitude as saturation.
Bottom: representative samples from real-world datasets (nuScenes and TruckScene).
}
\label{fig:dataset_vis}
\end{figure}

\paragraph{\textbf{Backbone Agnosticism.}}
To verify that the benefits of SynFlow dataset are not architecture-specific, we evaluate zero-shot performance across multiple scene flow backbones on the Aeva dataset (Tab.~\ref{tab:aeva_backbone_agnostic}). 
As an initial reference, an Ego Motion Flow baseline (assigning flow via odometry only) results in maximum Dynamic Normalized-Bucket EPE (1.000), demonstrating that ego-compensation alone cannot account for the dynamic scene elements. 
In contrast, after pre-training on SynFlow-4k, all learned backbones, including feed-forward estimators like DeFlow~\cite{zhang2024deflow} and Flow4D~\cite{kim2024flow4d}, as well as our default $\Delta$Flow, reduce both Dynamic Bucket-Normalized error and Three-way EPE. 
The performance gains across these diverse architectures indicate that SynFlow dataset provides a transferable and backbone-agnostic motion prior, validating its utility as a general-purpose supervisory source for LiDAR scene flow.

\subsection{Qualitative Comparison}
\label{sec:qualitative_com}
Fig.~\ref{fig:dataset_vis} visualizes representative samples from SynFlow-4k and real-world datasets. While real-world datasets provide essential sensor realism (patterns and noise), they are often tailored to specific operational domains: nuScenes focuses on urban driving with a 32-beam configuration, while TruckScenes targets highway environments with commercial vehicles. SynFlow-4k complements these specialized distributions by incorporating a variety of road topologies, including roundabouts and merging zones, across both 32 and 64-beam sensor profiles. It intentionally increases the density and diversity of dynamic agents, particularly pedestrians and small movers in sidewalks and crossing areas. Our pipeline also provides dense, point-wise supervision for all dynamic agents simultaneously, capturing kinematic details that can be difficult to annotate in real-world logs. 
Together, these properties allow the SynFlow dataset to serve as a comprehensive motion prior that complements the scene-specific coverage of individual real-world datasets.

\subsection{Limitations and Future Works}
\paragraph{\textbf{From Open-loop to Feedback-driven Synthesis.}}
Our current pipeline follows a ``generate-then-train'' paradigm: data is synthesized based on predefined policies and then used for training. This open-loop process does not dynamically adapt to the model's learning state.
Future work could explore a \emph{closed-loop} or \emph{cascade learning} framework, where the model's failure cases (e.g., specific occlusion patterns or rare motion speeds) act as feedback signals to trigger targeted re-simulation.
By actively generating ``hard examples'' adversarial to the current model, the pipeline could achieve higher data efficiency and continuous improvement.

\paragraph{\textbf{Expanding Domains.}}
While the presented methods and experiments focus on autonomous driving, the motion-oriented synthesis principle underlying SynFlow can generalize naturally to other domains where 3D motion labels are scarce. A natural extension is to apply this methodology to other simulation environments (e.g., Isaac Sim~\cite{isaac_sim_5_1_0}) for embodied robotics~\cite{Huang_2026_CVPR,zhang2026molmomotion}, covering indoor navigation, tabletop manipulation, and human-robot interaction scenarios.

\section{Conclusion}
\label{sec:conclusion}

In this work, we introduced SynFlow, a motion-oriented generation pipeline, and its resulting large-scale dataset, SynFlow-4k, to address the critical bottleneck of dense annotation in LiDAR scene flow. Rather than chasing perfect sensor realism, our approach prioritizes geometric and temporal multi-agent complexity, showing that physically consistent motion relations are highly transferable across domains.

Through extensive evaluations, SynFlow-4k is a robust, backbone-agnostic pre-training source. In a zero-shot regime, models trained on it generalize across diverse benchmarks, comparable to in-domain supervised performance. 
Fine-tuned on 5\% of real labels, it surpasses supervised baselines trained from scratch on four times the budget. As a complement to real-world data, it further supplies kinematic density for long-tail interactions that real-world logs lack. 
SynFlow provides a scalable, label-efficient data engine for generalizable dynamic 3D scene understanding.
We hope that releasing SynFlow and its extensible data engine will facilitate further research on generalizable 3D motion estimation and accelerate progress toward reliable dynamic perception in diverse real-world environments.

\section*{Acknowledgements}
This work was partially supported by the Wallenberg AI, Autonomous Systems and Software Program (WASP) funded by the Knut and Alice Wallenberg Foundation. 
The computations and data handling were enabled by the Berzelius supercomputing resource provided by the National Supercomputer Centre at Linköping University and the Knut and Alice Wallenberg Foundation, Sweden, as well as by resources provided by Chalmers e-Commons at Chalmers and the National Academic Infrastructure for Supercomputing in Sweden (NAISS), partially funded by the Swedish Research Council through grant agreement no. 2022-06725.

\bibliographystyle{splncs04}
\bibliography{main}

@String(CVPR  = {IEEE Conf. Comput. Vis. Pattern Recog.})

@String(ECCV  = {Eur. Conf. Comput. Vis.})

@String(NeurIPS = {Adv. Neural Inform. Process. Syst.})

@String(ICLR  = {Int. Conf. Learn. Represent.})

@String(CVPR  = {CVPR})

@String(ECCV  = {ECCV})

@String(NeurIPS = {NeurIPS})

@String(ICLR  = {ICLR})

@article{zhang2024gmsf,
  title={{GMSF}: Global matching scene flow},
  author={Zhang, Yushan and Edstedt, Johan and Wandt, Bastian and Forss{\'e}n, Per-Erik and Magnusson, Maria and Felsberg, Michael},
  journal={Advances in Neural Information Processing Systems},
  volume={36},
  year={2024}
}

@inproceedings{khatri2024can,
  title={I can’t believe it’s not scene flow!},
  author={Khatri, Ishan and Vedder, Kyle and Peri, Neehar and Ramanan, Deva and Hays, James},
  booktitle={European Conference on Computer Vision},
  pages={242--257},
  year={2024},
  organization={Springer}
}

@inproceedings{
zhang2025deltaflow,
title={{DeltaFlow}: An Efficient Multi-frame Scene Flow Estimation Method},
author={Zhang, Qingwen and Zhu, Xiaomeng and Zhang, Yushan and Cai, Yixi and Andersson, Olov and Jensfelt, Patric},
booktitle={The Thirty-ninth Annual Conference on Neural Information Processing Systems},
year={2025},
}

@inproceedings{hoffmann2025floxels,
  title={Floxels: Fast Unsupervised Voxel Based Scene Flow Estimation},
  author={Hoffmann, David T and Raza, Syed Haseeb and Jiang, Hanqiu and Tananaev, Denis and Klingenhoefer, Steffen and Meinke, Martin},
  booktitle={Proceedings of the Computer Vision and Pattern Recognition Conference},
  pages={22328--22337},
  year={2025}
}

@inproceedings{cai2023analyzing,
  title={Analyzing Infrastructure LiDAR Placement with Realistic LiDAR Simulation Library},
  author={Cai, Xinyu and Jiang, Wentao and Xu, Runsheng and Zhao, Wenquan and Ma, Jiaqi and Liu, Si and Li, Yikang},
  booktitle={2023 IEEE International Conference on Robotics and Automation (ICRA)},
  pages={5581--5587},
  year={2023},
  organization={IEEE}
}

@article{fent2024man,
  title={Man truckscenes: A multimodal dataset for autonomous trucking in diverse conditions},
  author={Fent, Felix and Kuttenreich, Fabian and Ruch, Florian and Rizwin, Farija and Juergens, Stefan and Lechermann, Lorenz and Nissler, Christian and Perl, Andrea and Voll, Ulrich and Yan, Min and others},
  journal={Advances in Neural Information Processing Systems},
  volume={37},
  pages={62062--62082},
  year={2024}
}

@ARTICLE{carlasc22,
  author={Wilson, Joey and Song, Jingyu and Fu, Yuewei and Zhang, Arthur and Capodieci, Andrew and Jayakumar, Paramsothy and Barton, Kira and Ghaffari, Maani},
  journal={IEEE Robotics and Automation Letters}, 
  title={MotionSC: Data Set and Network for Real-Time Semantic Mapping in Dynamic Environments}, 
  year={2022},
  volume={7},
  number={3},
  pages={8439-8446}}

@inproceedings{wang2026vggtomega,
  title     = {{VGGT-$\Omega$}},
  author    = {Jianyuan Wang and Minghao Chen and Shangzhan Zhang and Nikita Karaev and Johannes Sch{\"o}nberger and Patrick Labatut and Piotr Bojanowski and David Novotny and Andrea Vedaldi and Christian Rupprecht},
  booktitle = {Proceedings of the IEEE/CVF Conference on Computer Vision and Pattern Recognition (CVPR)},
  year      = {2026}
}

@inproceedings{jiang2023optimizing,
  title={Optimizing the Placement of Roadside LiDARs for Autonomous Driving},
  author={Jiang, Wentao and Xiang, Hao and Cai, Xinyu and Xu, Runsheng and Ma, Jiaqi and Li, Yikang and Lee, Gim Hee and Liu, Si},
  booktitle={Proceedings of the IEEE/CVF International Conference on Computer Vision},
  pages={18381--18390},
  year={2023}
}

@software{The_HDF_Group_Hierarchical_Data_Format,
author = {{The HDF Group}},
title = {{Hierarchical Data Format, version 5}},
note = {Accessed: 2026-06-28},
url = {https://github.com/HDFGroup/hdf5}
}

@article{li2025uniflowzeroshotlidarscene,
  title={{UniFlow}: Towards Zero-Shot LiDAR Scene Flow for Autonomous Vehicles via Cross-Domain Generalization},
  author={Li, Siyi and Zhang, Qingwen and Khatri, Ishan and Vedder, Kyle and Ramanan, Deva and Peri, Neehar},
  journal={arXiv preprint arXiv:2511.18254},
  year={2025}
}

@inproceedings{choy20194d,
  title={4D Spatio-Temporal ConvNets: Minkowski Convolutional Neural Networks},
  author={Choy, Christopher and Gwak, JunYoung and Savarese, Silvio},
  booktitle={Proceedings of the IEEE Conference on Computer Vision and Pattern Recognition},
  pages={3075--3084},
  year={2019}
}

@inproceedings{lin2025voteflow,
  title={VoteFlow: Enforcing Local Rigidity in Self-Supervised Scene Flow},
  author={Lin, Yancong and Wang, Shiming and Nan, Liangliang and Kooij, Julian and Caesar, Holger},
  booktitle={Proceedings of the Computer Vision and Pattern Recognition Conference},
  pages={17155--17164},
  year={2025}
}

@INPROCEEDINGS {Argoverse2_2021,
  author = {Benjamin Wilson and William Qi and Tanmay Agarwal and John Lambert and Jagjeet Singh and et al.},
  title = {Argoverse 2: Next Generation Datasets for Self-driving Perception and Forecasting},
  booktitle = {Proceedings of the Neural Information Processing Systems Track on Datasets and Benchmarks (NeurIPS Datasets and Benchmarks 2021)},
  year = {2021}
}

@inproceedings{zhang2024deflow,
  author={Zhang, Qingwen and Yang, Yi and Fang, Heng and Geng, Ruoyu and Jensfelt, Patric},
  booktitle={2024 IEEE International Conference on Robotics and Automation (ICRA)}, 
  title={{DeFlow}: Decoder of Scene Flow Network in Autonomous Driving}, 
  year={2024},
  pages={2105-2111},
  doi={10.1109/ICRA57147.2024.10610278}
}

@article{zeroflow,
    author    = {Kyle Vedder and Neehar Peri and Nathaniel Chodosh and Ishan Khatri and Eric Eaton and Dinesh Jayaraman and Yang Liu Deva Ramanan and James Hays},
    title     = {{ZeroFlow: Fast Zero Label Scene Flow via Distillation}},
    journal   = {International Conference on Learning Representations (ICLR)},
    year      = {2024},
}

@INPROCEEDINGS{nuscenes,
  title={nuScenes: A multimodal dataset for autonomous driving},
  author={Holger Caesar and Varun Bankiti and Alex H. Lang and Sourabh Vora and 
          Venice Erin Liong and Qiang Xu and Anush Krishnan and Yu Pan and 
          Giancarlo Baldan and Oscar Beijbom}, 
  booktitle={CVPR},
  year=2020
}

@inproceedings{sun2020scalability,
  title={Scalability in perception for autonomous driving: Waymo open dataset},
  author={Sun, Pei and Kretzschmar, Henrik and Dotiwalla, Xerxes and Chouard, Aurelien and Patnaik, Vijaysai and Tsui, Paul and Guo, James and Zhou, Yin and Chai, Yuning and Caine, Benjamin and others},
  booktitle={Proceedings of the IEEE/CVF conference on computer vision and pattern recognition},
  pages={2446--2454},
  year={2020}
}

@inproceedings{flythings3d,
  title={A large dataset to train convolutional networks for disparity, optical flow, and scene flow estimation},
  author={Mayer, Nikolaus and Ilg, Eddy and Hausser, Philip and Fischer, Philipp and Cremers, Daniel and Dosovitskiy, Alexey and Brox, Thomas},
  booktitle={Proceedings of the IEEE conference on computer vision and pattern recognition},
  pages={4040--4048},
  year={2016}
}

@INPROCEEDINGS{yangrealistic,
  author={Yang, Donglin and Cai, Xinyu and Liu, Zhenfeng and Jiang, Wentao and Zhang, Bo and Yan, Guohang and Gao, Xing and Liu, Si and Shi, Botian},
  booktitle={2024 IEEE/RSJ International Conference on Intelligent Robots and Systems (IROS)}, 
  title={Realistic Rainy Weather Simulation for LiDARs in CARLA Simulator}, 
  year={2024},
  volume={},
  number={},
  pages={951-957}}

@software{isaac_sim_5_1_0,
  author       = {{NVIDIA}},
  title        = {{Isaac Sim}},
  version      = {5.1.0},
  year         = {2024},
  url          = {https://github.com/isaac-sim/IsaacSim},
  note = {Accessed: 2026-06-28},
  license      = {Apache-2.0}
}

@inproceedings{yang2023vidar,
  title={3DSFLabelling: Boosting 3D Scene Flow Estimation by Pseudo Auto-labelling},
  author={Jiang, Chaokang and Wang, Guangming and Liu, Jiuming and Wang, Hesheng and Ma, Zhuang and Liu, Zhenqiang and Liang, Zhujin and Shan, Yi and Du, Dalong},
  booktitle={Proceedings of the IEEE/CVF Conference on Computer Vision and Pattern Recognition},
  year={2024}
}

@inproceedings{pang2022simpletrack,
  title={Simpletrack: Understanding and rethinking 3d multi-object tracking},
  author={Pang, Ziqi and Li, Zhichao and Wang, Naiyan},
  booktitle={European conference on computer vision},
  pages={680--696},
  year={2022},
  organization={Springer}
}

@article{wang2023technical,
  title={Technical report for argoverse challenges on unified sensor-based detection, tracking, and forecasting},
  author={Wang, Zhepeng and Chen, Feng and Lertniphonphan, Kanokphan and Chen, Siwei and Bao, Jinyao and Zheng, Pengfei and Zhang, Jinbao and Huang, Kaer and Zhang, Tao},
  journal={arXiv preprint arXiv:2311.15615},
  year={2023}
}

@misc{aevascenes,
    title        = {AevaScenes: A Dataset and Benchmark for FMCW LiDAR Perception},
    author       = {Narasimhan, Gautham Narayan and Vhavle, Heethesh and Vishvanatha, Kumar Bhargav and Reuther, James},
    year         = {2025},
    url          = {https://scenes.aeva.com/},
    note = {Accessed: 2026-06-28}
  }

@inproceedings{zhang2026teflow,
  title = {{TeFlow}: Enabling Multi-frame Supervision for Self-Supervised Feed-forward Scene Flow Estimation},
  author={Zhang, Qingwen and Jiang, Chenhan and Zhu, Xiaomeng and Miao, Yunqi and Zhang, Yushan and Andersson, Olov and Jensfelt, Patric},
  year = {2026},
  booktitle = {Proceedings of the IEEE/CVF conference on computer vision and pattern recognition},
  pages = {3667-3676},
}

@inproceedings{sun2022shift,
  title={SHIFT: a synthetic driving dataset for continuous multi-task domain adaptation},
  author={Sun, Tao and Segu, Mattia and Postels, Janis and Wang, Yuxuan and Van Gool, Luc and Schiele, Bernt and Tombari, Federico and Yu, Fisher},
  booktitle={Proceedings of the IEEE/CVF conference on computer vision and pattern recognition},
  pages={21371--21382},
  year={2022}
}

@inproceedings{
zhang2024resimad,
title={ReSim{AD}: Zero-Shot 3D Domain Transfer for Autonomous Driving with Source Reconstruction and Target Simulation},
author={Bo Zhang and Xinyu Cai and Jiakang Yuan and Donglin Yang and Jianfei Guo and Xiangchao Yan and Renqiu Xia and Botian Shi and Min Dou and Tao Chen and Si Liu and Junchi Yan and Yu Qiao},
booktitle={The Twelfth International Conference on Learning Representations},
year={2024}
}

@inproceedings{kloukiniotis2022carlascenes,
  title={Carlascenes: A synthetic dataset for odometry in autonomous driving},
  author={Kloukiniotis, Andreas and Papandreou, Andreas and Anagnostopoulos, Christos and Lalos, Aris and Kapsalas, Petros and Nguyen, Duong-Van and Moustakas, Konstantinos},
  booktitle={Proceedings of the IEEE/CVF Conference on Computer Vision and Pattern Recognition},
  pages={4520--4528},
  year={2022}
}

@inproceedings{dosovitskiy2017carla,
  title={CARLA: An open urban driving simulator},
  author={Dosovitskiy, Alexey and Ros, German and Codevilla, Felipe and Lopez, Antonio and Koltun, Vladlen},
  booktitle={Conference on robot learning},
  pages={1--16},
  year={2017},
  organization={PMLR}
}

@inproceedings{vanhoorick2024gcd,
title={Generative Camera Dolly: Extreme Monocular Dynamic Novel View Synthesis},
author={Van Hoorick, Basile and Wu, Rundi and Ozguroglu, Ege and Sargent, Kyle and Liu, Ruoshi and Tokmakov, Pavel and Dave, Achal and Zheng, Changxi and Vondrick, Carl},
booktitle=ECCV,
year={2024}
}

@inproceedings{ren2025gen3c,
    title={GEN3C: 3D-Informed World-Consistent Video Generation with Precise Camera Control},
    author={Ren, Xuanchi and Shen, Tianchang and Huang, Jiahui and Ling, Huan and
        Lu, Yifan and Nimier-David, Merlin and Müller, Thomas and Keller, Alexander and Fidler, Sanja and Gao, Jun},
    booktitle=CVPR,
    year={2025}
}

@inproceedings{vggt,
  author       = {Jianyuan Wang and
                  Minghao Chen and
                  Nikita Karaev and
                  Andrea Vedaldi and
                  Christian Rupprecht and
                  David Novotn{\'{y}}},
  title        = {{VGGT:} Visual Geometry Grounded Transformer},
  booktitle    = {{IEEE/CVF} Conference on Computer Vision and Pattern Recognition},
  pages        = {5294--5306},
  year         = {2025},
}

@inproceedings{jiang2025megasynth,
  title={Megasynth: Scaling up 3d scene reconstruction with synthesized data},
  author={Jiang, Hanwen and Xu, Zexiang and Xie, Desai and Chen, Ziwen and Jin, Haian and Luan, Fujun and Shu, Zhixin and Zhang, Kai and Bi, Sai and Sun, Xin and others},
  booktitle={Proceedings of the Computer Vision and Pattern Recognition Conference},
  pages={16441--16452},
  year={2025}
}

@article{xie2024lrm,
  title={Lrm-zero: Training large reconstruction models with synthesized data},
  author={Xie, Desai and Bi, Sai and Shu, Zhixin and Zhang, Kai and Xu, Zexiang and Zhou, Yi and Pirk, S{\"o}ren and Kaufman, Arie and Sun, Xin and Tan, Hao},
  journal={Advances in Neural Information Processing Systems},
  volume={37},
  pages={53285--53316},
  year={2024}
}

@INPROCEEDINGS{chodosh2023re,
  author={Chodosh, Nathaniel and Ramanan, Deva and Lucey, Simon},
  booktitle={2024 IEEE/CVF Winter Conference on Applications of Computer Vision (WACV)}, 
  title={Re-Evaluating LiDAR Scene Flow}, 
  year={2024},
  pages={5993-6003},
  doi={10.1109/WACV57701.2024.00590}
}

@inproceedings{lin2024icp,
  title={{ICP-Flow}: LiDAR Scene Flow Estimation with ICP},
  author={Lin, Yancong and Caesar, Holger},
  booktitle={CVPR},
  year={2024}
}

@article{zhang2025himo,
  title={{HiMo}: High-Speed Objects Motion Compensation in Point Cloud},
  author={Zhang, Qingwen and Khoche, Ajinkya and Yang, Yi and Ling, Li and Mansouri, Sina Sharif and Andersson, Olov and Jensfelt, Patric},
  journal={IEEE Transactions on Robotics}, 
  year={2025},
  volume={41},
  number={},
  pages={5896-5911},
  doi={10.1109/TRO.2025.3619042}
}

@inproceedings{alibeigi2023zenseact,
      title={Zenseact Open Dataset: A large-scale and diverse multimodal dataset for autonomous driving},
      author={Alibeigi, Mina and Ljungbergh, William and Tonderski, Adam and Hess, Georg and Lilja, Adam and Lindstrom, Carl and Motorniuk, Daria and Fu, Junsheng and Widahl, Jenny and Petersson, Christoffer},
      booktitle={Proceedings of the IEEE/CVF International Conference on Computer Vision},
      year={2023}
}

@inproceedings{once21,
  author={Jiageng Mao and Minzhe Niu and Chenhan Jiang and Hanxue Liang and Jingheng Chen and Xiaodan Liang and Yamin Li and Chaoqiang Ye and Wei Zhang and Zhenguo Li and Jie Yu and Chunjing Xu and Hang Xu},
  title={One Million Scenes for Autonomous Driving: ONCE Dataset},
  year={2021},
  booktitle={NeurIPS Datasets and Benchmarks}
}

@article{mambaflow,
  author={Luo, Jiehao and Cheng, Jintao and Zhang, Qingwen and Xue, Bohuan and Fan, Rui and Tang, Xiaoyu},
  journal={IEEE Transactions on Intelligent Vehicles}, 
  title={{MambaFlow}: A Novel and Flow-Guided State Space Model for Scene Flow Estimation}, 
  year={2026},
  volume={11},
  number={4},
  pages={511-521},
  doi={10.1109/TIV.2026.3663171}
}

@inproceedings{zhang2024seflow,
  author={Zhang, Qingwen and Yang, Yi and Li, Peizheng and Andersson, Olov and Jensfelt, Patric},
  title={{SeFlow}: A Self-Supervised Scene Flow Method in Autonomous Driving},
  booktitle={European Conference on Computer Vision (ECCV)},
  year={2024},
  pages={353–369},
  organization={Springer},
  doi={10.1007/978-3-031-73232-4_20},
}

@article{kim2024flow4d,
  author={Kim, Jaeyeul and Woo, Jungwan and Shin, Ukcheol and Oh, Jean and Im, Sunghoon},
  journal={IEEE Robotics and Automation Letters}, 
  title={{Flow4D}: Leveraging 4D Voxel Network for LiDAR Scene Flow Estimation}, 
  year={2025},
  volume={},
  number={},
  pages={1-8},
  doi={10.1109/LRA.2025.3542327}
}

@inproceedings{liu2024difflow3d,
  title={DifFlow3D: Toward Robust Uncertainty-Aware Scene Flow Estimation with Iterative Diffusion-Based Refinement},
  author={Liu, Jiuming and Wang, Guangming and Ye, Weicai and Jiang, Chaokang and Han, Jinru and Liu, Zhe and Zhang, Guofeng and Du, Dalong and Wang, Hesheng},
  booktitle={Proceedings of the IEEE/CVF Conference on Computer Vision and Pattern Recognition},
  pages={15109--15119},
  year={2024}
}

@inproceedings{vedder2024neural,
title={Neural Eulerian Scene Flow Fields},
author={Kyle Vedder and Neehar Peri and Ishan Khatri and Siyi Li and Eric Eaton and Mehmet Kemal Kocamaz and Yue Wang and Zhiding Yu and Deva Ramanan and Joachim Pehserl},
booktitle={The Thirteenth International Conference on Learning Representations},
year={2025},
url={https://openreview.net/forum?id=0CieWy9ONY}
}

@InProceedings{Huang_2026_CVPR,
    author    = {Huang, Wenlong and Chao, Yu-Wei and Mousavian, Arsalan and Liu, Ming-Yu and Fox, Dieter and Mo, Kaichun and Fei-Fei, Li},
    title     = {PointWorld: Scaling 3D World Models for In-The-Wild Robotic Manipulation},
    booktitle = {Proceedings of the IEEE/CVF Conference on Computer Vision and Pattern Recognition (CVPR)},
    month     = {June},
    year      = {2026},
    pages     = {20765-20779}
}

@article{zhang2026molmomotion,
    title         = {MolmoMotion: Forecasting Point Trajectories in 3D with Language Instruction},
    author        = {Zhang, Jianing and Zheng, Chenhao and Yang, Yajun and Argus, Max and Soraki, Rustin and Han, Winson and Anderson, Taira and Li, Chun-Liang and Liu, Shuo and Duan, Jiafei and Ren, Zhongzheng and Zhang, Jieyu and Krishna, Ranjay},
    journal       = {arXiv preprint arXiv:2606.18558},
    year          = {2026},
}
\end{document}